\documentclass{article}

\usepackage{microtype}
\usepackage{graphicx}
\usepackage{subfigure}
\usepackage{booktabs} 

\usepackage{hyperref}

\usepackage[accepted]{icml2024}

\usepackage{amsmath}
\usepackage{amssymb}
\usepackage{mathtools}
\usepackage{amsthm}

\usepackage[capitalize,noabbrev]{cleveref}

\theoremstyle{plain}

\theoremstyle{definition}

\theoremstyle{remark}

\usepackage[textsize=tiny]{todonotes}

\icmltitlerunning{Post-hoc Part-prototype Networks}

\begin{document}

\twocolumn[
\icmltitle{Post-hoc Part-prototype Networks}



\icmlsetsymbol{equal}{*}

\begin{icmlauthorlist}
\icmlauthor{Andong Tan}{yyy}
\icmlauthor{Fengtao Zhou}{yyy}
\icmlauthor{Hao Chen}{yyy,aging}
\end{icmlauthorlist}

\icmlaffiliation{yyy}{Department of Computer Science and Engineering, Hong Kong University of Science and Technology, Hong Kong, China}
\icmlaffiliation{aging}{Center for Aging Science, HKUST, Hong Kong, China}

\icmlcorrespondingauthor{Hao Chen}{jhc@cse.ust.hk}

\icmlkeywords{Machine Learning, ICML}

\vskip 0.3in
]



\printAffiliationsAndNotice{}  

\begin{abstract}
Post-hoc explainability methods such as Grad-CAM are popular because they do not influence the performance of a trained model. However, they mainly reveal ``where'' a model looks at for a given input, fail to explain ``what'' the model looks for (e.g., what is important to classify a bird image to a Scott Oriole?). Existing part-prototype networks leverage part-prototypes (e.g., characteristic Scott Oriole's wing and head) to answer both ``where" and ``what", but often under-perform their black box counterparts in the accuracy. Therefore, a natural question is: can one construct a network that answers both ``where'' and ``what" in a post-hoc manner to guarantee the model's performance? To this end, we propose the first post-hoc part-prototype network via decomposing the classification head of a trained model into a set of interpretable part-prototypes. Concretely, we propose an unsupervised prototype discovery and refining strategy to obtain prototypes that can precisely reconstruct the classification head, yet being interpretable. Besides guaranteeing the performance, we show that our network offers more faithful explanations qualitatively and yields even better part-prototypes quantitatively than prior part-prototype networks.
\end{abstract}

\section{Introduction}
In the past years, deep learning based models such as Convolutional Neural Networks (CNN) and Vision Transformer (ViT) have achieved impressive performance in computer vision tasks \citep{he2016deep, dosovitskiy2020image, girshick2015fast, ronneberger2015u}. However, the interpretability of these black-box models has always been a major concern and limits their real-world deployments.

\begin{figure}
    \centering
    \includegraphics[width=\linewidth]{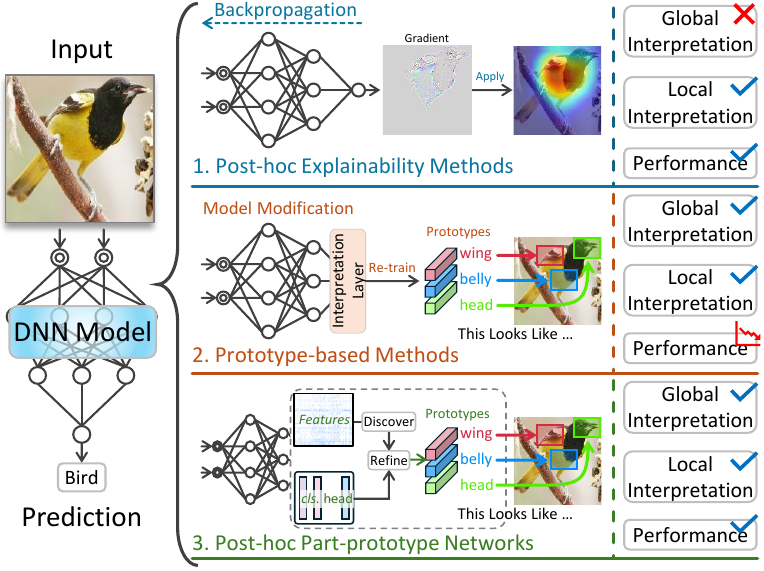}
    \caption{Prior post-hoc methods such as Grad-CAM \cite{selvaraju2017grad} fall short in answering ``what the model looks for" (a type of global interpretability) for transparent decision making. On the other hand, prior part-prototype based models often sacrifice the performance. Our post-hoc part-prototype network is the first model offering part-prototype based global interpretability for decision making, while guaranteeing the prediction performance.}
    \vspace{-0.5cm}
    \label{fig:teaser}
\end{figure}

Existing post-hoc methods such as Grad-CAM \cite{selvaraju2017grad} can explain a trained model (the name post-hoc comes from explaining a model after it is trained) by generating a heatmap indicating where the model looks at. These methods are popular because they do not influence the model's performance. However, they have been shown to fail, or only marginally help in recent human-centered psychophysics experiments \cite{colin2022cannot, kim2022hive}, mainly due to their failures to indicate what the model is looking for to make certain decisions \cite{fel2023craft}. As a concrete example, given an image of a bird, Grad-CAM \cite{selvaraju2017grad} only indicates that a model is looking at some foreground areas of the input (a type of local interpretability as the explanation is locally valid to a certain input), but fails to tell what visual features in these areas are driving the decision \cite{rudin2019stop, chen2019looks}. Therefore, they can not explain ``what is generally important for the model to classify a bird into a Scott Oriole" (a type of global interpretability as the explanation will be globally valid in the whole input space) \cite{schwalbe2023comprehensive}.

To address these concerns, part-prototype networks are recently proposed. They leverage the case-based reasoning framework to point out ``this looks like that" \cite{chen2019looks, nauta2021neural, wang2021interpretable, rymarczyk2022interpretable, donnelly2022deformable,huang2023evaluation}. These models classify images by comparing inputs with a set of prototypical parts and can thus tell what the network is looking for to make certain decisions. Since this reasoning process is transparent and the contribution from each prototype is clear, these models are also considered self-explainable \cite{alvarez2018towards}. However, existing methods require model modifications (e.g., adding additional prototype layers) and re-training the whole network with intricate loss function designs, making them less convenient to be applied. Moreover, these models often under-perform their black box counterparts in the accuracy. A conceptual comparison of pros and cons of different approaches is summarized in Figure \ref{fig:teaser}.

Motivated by the above observation, a natural question is: \textbf{can we construct a model that explains both where the model looks at and what the model looks for in a post-hoc manner to guarantee the performance?} To this end, we propose to directly decompose a trained classification head into a set of interpretable part-prototypes. In such a manner, the model will be capable to explain ``this bird is classified to Scott Oriole because this area looks like a Scott Oriole's head and that area looks like a Scott Oriole's wing" instead of ``it is classified to the Scott Oriole because of these areas", which does not explain what visual features the model is looking for in these areas. To enable such an interpretable decomposition, we propose a novel part-prototype discovery and refining strategy. Concretely, we discover initial part-prototypes via applying None-negative Matrix Factorization (NMF) on the features due to its known capabilities to yield parts based representations \cite{lee1999learning}. Since these initial part-prototypes may not be able to fully reconstruct a trained high dimensional classification head, we further propose a prototype refinement step via an optimization based residual parameter distribution subject to interpretability constraints. The distributed parameters move the interpretable prototypes to a more class-discriminative, yet semantically meaningful areas of the feature space to fully recover the model's performance.

Our contributions can be summarized as follows:
\begin{itemize}
    \item We propose the first post-hoc part-prototype network, which can explain both where a trained network looks at and what it looks for without model modifications or re-training, while guaranteeing the performance.
    \item We propose an unsupervised prototype discovery and refining strategy to precisely reconstruct a trained model's classification head with no approximation error, yet maintaining the interpretability.
    \item We conduct extensive qualitative and quantitative evaluations using various backbones and datasets to show both interpretability and performance advantages over prior part-prototype networks. 
\end{itemize}

\section{Related Work}
\paragraph{Part-prototype models.}
One prominent work designing part-prototype networks for interpretable image classification is the ProtopNet \citep{chen2019looks}. This work focuses on offering a transparent reasoning process: given an input image, the model compares the image features with the prototypes, and make the predictions based on a weighted combination of the similarity scores between the image and part-prototypes (e.g., a part of the object with semantic meaning from a certain class). This model provides inherent interpretability of the decision making in a ``this looks like that" manner, answering what the model generally looks for to make certain decisions. Following this framework, many works are proposed to investigate different aspects, such as discovering the similarities of prototypes \citep{rymarczyk2021protopshare}, making the prototype's class assignment differentiable \citep{rymarczyk2022interpretable}, 
making the prototypes spatially flexible \citep{donnelly2022deformable}, 
or illuminating the prototypes with multiple visualizations for richer explanations \cite{ma2023looks},
increasing the prototype stability \cite{huang2023evaluation},
combining it with decision trees \citep{nauta2021neural} or K-nearest neighbors (KNN) \citep{ukai2022looks}. These methods require model modifications (e.g., adding additional prototype layers), re-training the whole network and are thus less convenient to be applied. More importantly, they often exhibit performance drops and the faithfulness of the explanations explaining ``where" are recently challenged \cite{wolf2023keep}. In contrast, our post-hoc part-prototype network guarantees the performance without additional network re-training and easily fulfills properties desired by a faithful explanation when explaining ``where" the model looks at.

\paragraph{Post-hoc methods.}
These methods focus on explaining trained models. Prior post-hoc methods can roughly be categorized to perturbation \citep{ribeiro2016should, zeiler2014visualizing, zhou2015predicting} based or backpropagation \citep{selvaraju2017grad, zhang2018top,sundararajan2017axiomatic,kapishnikov2021guided, yang2023idgi} based. These methods primarily emphasize local interpretability, providing explanations specific to individual inputs (e.g., explaining ``where"). However, they often fall short in terms of capturing global knowledge and utilizing it for transparent reasoning processes (e.g., explaining ``what") \cite{chen2019looks}. In contrast, our method equips the model the capability to explain what certain areas of the feature map look similar to, offering a part-prototype based global interpretability.

\begin{figure*}[th!]
    \centering
    \includegraphics[width=0.9\linewidth]{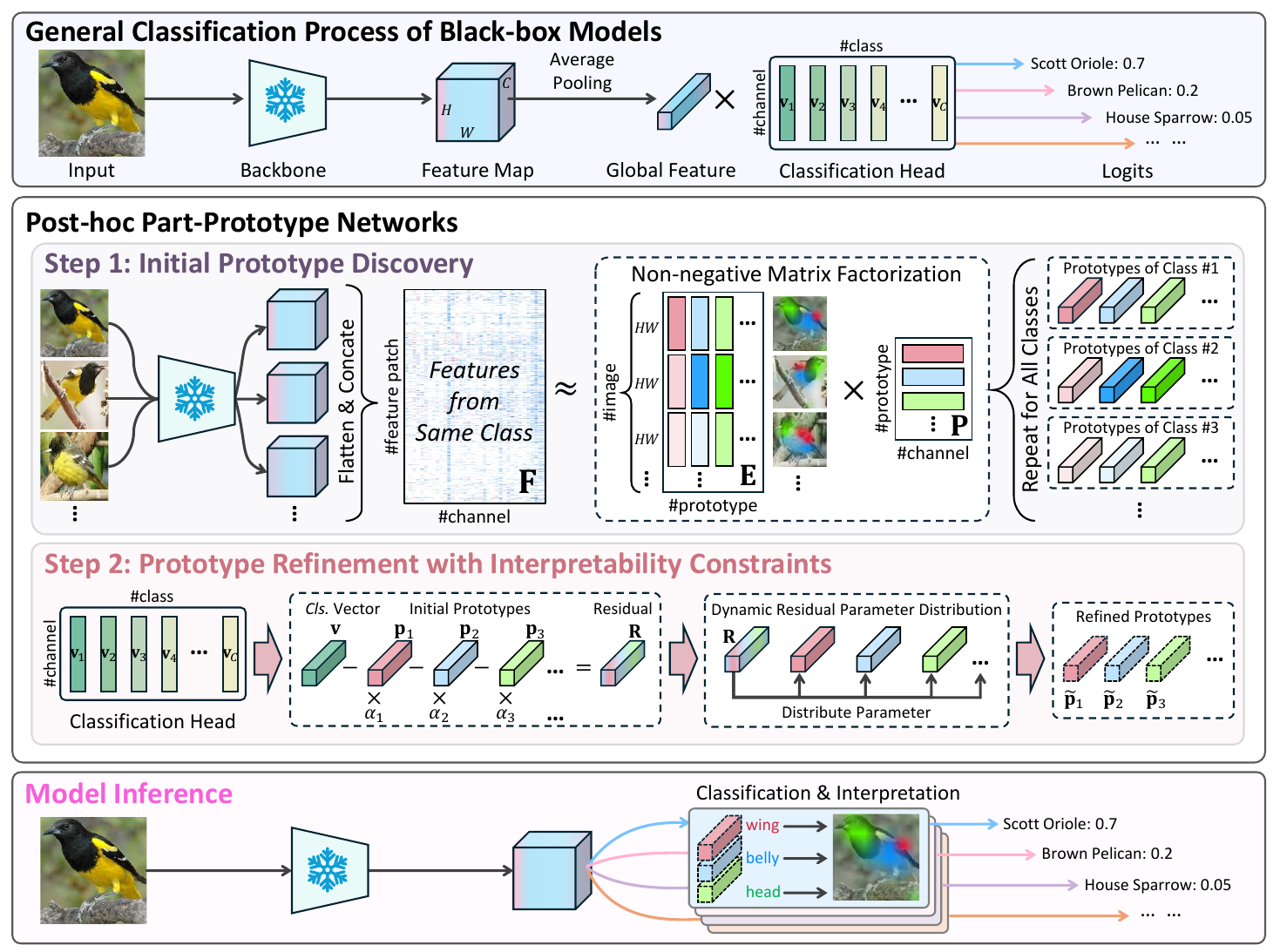}
    \caption{Overview of constructing a Post-hoc Part-Prototype Network. Given a trained black box model, we aim to fully decompose the classification head into interpretable prototypes. To achieve this, we first discover a set of prototypes $\mathbf{P}$ via NMF (step 1) for each class. In each image, we display the heatmaps of 3 prototypes in different colors for easier comparison. Visualizations are created via up-sampling the feature map to the original image resolution for ease of reading. Then we refine these prototypes $\mathbf{P}$ via scaling and dynamic residual parameter distribution subject to interpretability constraints. This step aims to guarantee a precise reconstruction only using part-prototypes without sacrificing the interpretability of them.}
    \vspace{-0.05cm}
    \label{fig:framework}
\end{figure*}

\paragraph{Matrix factorization and basis decomposition.}
It is not new to understand complex signals via the decomposition. Conventional methods such as NMF \citep{lee1999learning}, Principle Component Analysis \citep{frey1978principal}, Vector Quantization \citep{gray1984vector}, Independent Component Analysis \citep{hyvarinen2000independent}, Bilinear Models \citep{tenenbaum1996separating} and Isomap \citep{tenenbaum2000global} all discover meaningful subspace from the raw data matrix in an unsupervised manner \citep{zhou2018interpretable}. Among these works, NMF \citep{lee1999learning} stands out as the only approach capable of decomposing whole images into parts based representations due to its use of non-negativity constraints which allow only additive, not subtractive combinations. Parts-based representations have significant interpretability benefits because psychological \citep{biederman1987recognition} and physiological \citep{wachsmuth1994recognition} study show that human recognize objects by components. In the era of deep learning, \citep{collins2018deep,fel2023craft} find that NMF is also effective for discovering interpretable concepts in CNN features with ReLU \citep{nair2010rectified} as activation functions. However, both works do not leverage the discovered concepts from the training data to construct a model with transparent reasoning process. 
Another related work is \citep{zhou2018interpretable}, which approximates the classification head vector via a set of interpretable basis vectors to facilitate interpretable decision making. However, this work leverages the concept annotations to obtain the interpretable basis rather than discovering them in an unsupervised manner. What's more, the interpretable basis in this work fail to precisely reconstruct the classification head. Therefore, this method can not guarantee the same performance as the original model.

To the best of our knowledge, this is the first work constructing a part-prototype model in a post-hoc manner, significantly simplifying the process of obtaining part-prototype based global interpretability while guaranteeing the same prediction performance as the black box model.

\section{Method}
In this section, we first introduce our designing goal and then elaborate how this goal could be achieved via our proposed prototype discovery and refinement strategy.

\subsection{Goal: Interpretable Classification Head Decomposition}
We first denote the feature map of an image extracted by the backbone as $\mathbf{x}\in \mathbb{R}^{HW\times D}$, where $H,W,D$ are the height, width and channel dimension respectively. The complete classification head is denoted by $\mathbf{V}= [\mathbf{v}^1, \mathbf{v}^2,...,\mathbf{v}^C] \in \mathbb{R}^{C\times D}$ for a model classifying $C$ classes. Without loss of generality, we consider one specific class $c$ in the following description and omit the class index in all notations for simplicity. Prior post-hoc class activation map could explain where the network looks at via $\mathbf{xv}^T \in \mathbb{R}^{HW\times 1}$, and the logit of classifying an image into class c is the average value of this matrix. This formulation, however, does not tell what more fine-grained components the model looks for to make the decision. This is desired because psychological \citep{biederman1987recognition} and physiological \citep{wachsmuth1994recognition} study show that human recognize objects by components. Therefore, we propose to decompose the classification vector into $k$ part-prototypes $\Tilde{\mathbf{p}}_i \in \mathbb{R}^{1\times D}$ as follows:
\begin{equation}
    \mathbf{v} = \Tilde{\mathbf{p}}_1+\Tilde{\mathbf{p}}_2 +...+\Tilde{\mathbf{p}}_k.
\end{equation}
This simple form can easily explain what important components the model looks for (e.g., $\Tilde{\mathbf{p}}_i$ represents a characteristic bird head, belly or wing) to classify an image into class $c$, and explain where the model looks at via visualizing $k$ heatmaps $\mathbf{x}\Tilde{\mathbf{p}}_i^T \in \mathbb{R}^{HW\times 1}$, as shown in the model inference step of Figure \ref{fig:framework} (we put 3 heatmaps in the same image with different colors for easier comparison between different prototypes). $k$ is a relatively small number because a human is typically not capable to decompose an object into too many components \cite{ramaswamy2023overlooked}. In the next section, we describe how to find such an interpretable decomposition. 

\subsection{Prototype Discovery and Refinement}
To find these prototypes, we propose to: (1) Discover a set of initial interpretable prototypes from image feature maps via None-negative Matrix Factorization. (2) Scale these prototypes to approximate the classification head via a convex optimization. (3) Refine these scaled prototypes via an optimization based residual parameter distribution subject to interpretability constraints.

Formally, given $n$ images from class $c$, the $i^{th}$ image's features extracted by neural networks are flattened as $\mathbf{f_i}\in \mathbb{R}^{HW\times D}$. The features from $n$ images stacked together are denoted as $\mathbf{F} = [\mathbf{f}_1, ...,\mathbf{f}_n] \in \mathbb{R}^{nHW\times D}$.

\paragraph{Initial prototype discovery.} This is achieved via applying None-negative Matrix Factorization (NMF) to optimize the following objective function:
\begin{equation}
\label{eqa:feature_err}
    \min_{\mathbf{E, P}} ||\mathbf{F}-\mathbf{EP}||_2^2,
\end{equation}
where $\mathbf{P} \in \mathbb{R}^{k\times D}$ are the $k$ initial prototypes we find and $\mathbf{E}\in \mathbb{R}^{nHW\times k}$ indicates how strong each prototype contributes to the reconstruction of features in all $HW$ spatial positions of all $n$ image feature maps.

The choice of using NMF is motivated by the fact that it only allows additive, and not subtractive combinations of components, removing complex cancellations of positive and negative values in $\mathbf{EP}$ \citep{lee1999learning}. Such a constraint on the sign of the decomposed matrices is proved to lead to sparsity \citep{lee2000algorithms} and also parts-based representations \citep{wang2012nonnegative}. It is therefore suitable to discover part-prototypes to indicate individual components (e.g., a Scott Oriole's head, wing, belly) of the target object (e.g., a Scott Oriole bird) under the part-prototype network framework. Popular matrix decomposition technique such as Principle Component Analysis does not have this capability, as shown in \cite{lee1999learning}. When applying NMF, we stop the optimization when the error difference between 2 updates with respect to the initial error is lower than $10^{-4}$ or 200 iterations are reached. 

\paragraph{Prototype scaling.} In this step, we first try to approximately reconstruct the trained classification head $\mathbf{v} \in \mathbb{R}^{1\times D}$ via initially discovered part-prototypes $\mathbf{P}$. We propose the following convex optimization:
\begin{equation}
\label{eqa:cls_err}
    \min_{\alpha_i} ||\mathbf{v} - \sum_{i=1}^{k}\alpha_i\mathbf{p}_i||_2^2,
\end{equation}
where $\alpha_i$ is the coefficient to scale the initial prototypes with respect to the class $c$ and $\mathbf{p}_i$ is a row vector of $\mathbf{P}$. 
Since $\mathbf{v}$ and $\mathbf{p}_i$ are fixed, this optimization is convex and $\alpha_i$ has a global optimum. This optimization tries to express the classification head (or target object) as a combination of initially discovered part-prototypes. However, due to the typically small number of part-prototypes $k$, there is no guarantee that a linear combination of the basis consisting of these $k$ initially discovered prototypes can precisely reconstruct the trained high-dimensional (e.g., 512 in ResNet34) classification head. 
Thus we further propose to refine these prototypes to fully recover the model performance.

\begin{figure}
    \centering
    \includegraphics[width=\linewidth]{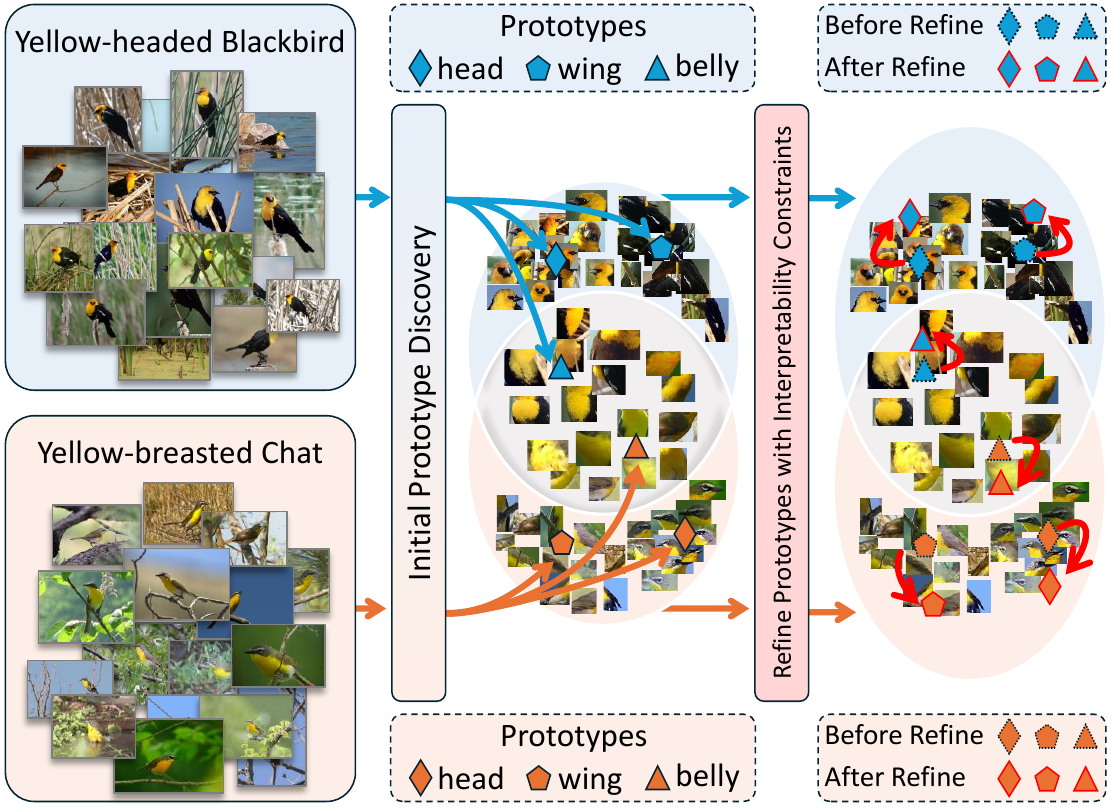}
    \vspace{-0.3cm}
    \caption{After discovering initial prototypes via None-negative Matrix Factorization, our refinement step makes the part-prototypes more discriminative for complete performance recovery, while maintaining their interpretability.}
    \label{fig:proto_refinement}
\end{figure}

\paragraph{Prototype refinement.} We propose to refine the prototypes via distributing the residual parameters.
This step aims to equip interpretable prototypes with better class-discriminative power and thus fully recover the model's performance (an illustrative example is offered in the Figure \ref{fig:proto_refinement}). After the coefficients $\alpha_i$ are already optimized, the residual parameters $\mathbf{R} \in \mathbb{R}^{1\times D}$ are calculated as:
\begin{equation}
    \mathbf{R} = \mathbf{v} - \sum_{i=1}^k \alpha_i \mathbf{p}_i.
\end{equation}
In order to completely decompose the classification head into $k$ interpretable prototypes, we suggest that the parameters in $\mathbf{R}$ should be fully absorbed by interpretable prototypes, while not sacrificing the prototypes' interpretability. Thus we propose to decompose the residual parameters into 
\begin{equation}
    \mathbf{R} = \mathbf{r}_1 + \mathbf{r}_2+...+ \mathbf{r}_k
\end{equation}
and distribute above $k$ components $\mathbf{r_i} \in \mathbb{R}^{1\times D}$ to refine all $k$ prototypes respectively. This means shifting the part-prototypes to a more class-discriminative, yet semantically meaningful areas in the feature space. A straightforward method to obtain $\mathbf{r_i}$ is setting $\mathbf{r}_i=\frac{a_i\mathbf{R}}{\sum_{i=1}^k \alpha_i}$, so $\sum_{i=1}^{k} \mathbf{r}_i=\mathbf{R}$ will automatically hold. However, such a naive strategy may harm the interpretability of original prototypes, because the $\mathbf{r}_i$ could shift the prototypes to some semantically less meaningful areas of the feature space. Therefore, we further propose the following dynamic optimization based prototype refinement strategy with interpretability constraints:
\begin{equation}
\begin{aligned}
\begin{gathered}
   \min_{\mathbf{r_i}} \sum_{i=1}^{k}||\mathrm{Norm}(\mathbf{F}\mathbf{p}_i^T)-\mathrm{Norm}(\mathbf{F}\mathbf{r}_i^T)||_2^2 \\
    s.t. \sum_{i=1}^k \mathbf{r}_i=\mathbf{R} .
\end{gathered}
\end{aligned}
\end{equation}
The $\mathrm{Norm}$ indicates the normalization operation along the spatial dimension of images (e.g., the first dimension of $\mathbf{Fp}_i^T\in \mathbb{R}^{nHW\times1}$). We try to align the distribution after the normalization because human perceives heatmap images according to the relative value distribution in the heatmaps to observe the highlighted areas. This constraint encourages the visualization maps generated by $\mathbf{r}_i$ with respect to $\mathbf{F}$ to be perceptually close to that of the $\mathbf{p}_i$ to maintain the interpretability of $\mathbf{p}_i$ after the refinement. In the implementation, we use the ``Nelder-Mead" \cite{nelder1965simplex} method with tolerance $10^{-6}$ and maximal iteration 100 to solve this optimization problem. 

After completing all above steps, the classification head for class $c$ is finally completely decomposed into $k$ interpretable prototypes $\mathbf{\Tilde{p_i}}$:
\begin{equation}
    \mathbf{v} = \sum_{i=1}^k \Tilde{\mathbf{p}}_i, \quad \Tilde{\mathbf{p}}_i = \alpha_i\mathbf{p}_i+\mathbf{r}_i.
\end{equation}
Repeating above steps for all classes could decompose the whole trained classification head of a model. 

\section{Experiments}
In this section, we first leverage commonly used explainability axioms (desired characteristics) in prior post-hoc methods mainly explaining ``where" to show these axioms are easily fulfilled by our method, while prior prototype based models fail in sensitivity, completeness and linearity axioms qualitatively. We call a method fulfilling these axioms faithful explanations following \cite{wolf2023keep}. Then we compare with existing part-prototype models in explaining ``what" the model looks for and demonstrate the benefits quantitatively that our method yields more consistent and stable prototypes, yet guaranteeing the accuracy in the CUB-200-2011 benchmark dataset \cite{wah2011caltech} in 5 backbones. In addition, we show our method can be easily applied to large scale datasets such as ImageNet \cite{deng2009imagenet}, while prior prototype based models either fail due to huge memory consumption or exhibit significant performance drops. 
Comprehensive ablation study is offered to discuss the importance and influence of our design choices.

\subsection{Comparisons via Explainability Axioms}
We show qualitatively that our method yields more faithful explanations than prior part-prototype models in explaining ``where" the model looks at via a set of explainability axioms.

Different works propose slightly different sets of axioms and we summarize the commonly desired properties that an explanation should fulfill when attributing the prediction of a model to the contribution of individual input features in this paragraph \cite{sundararajan2017axiomatic,lundberg2017unified}. \textbf{Sensitivity:} The attribution should be zero if the prediction does not depend on a certain feature and non-zero otherwise.
\textbf{Completeness:} The sum of contributions of all features equals the prediction.
\textbf{Linearity:} If a model is a linear combination of multiple models, the attributions of it should have the same linearity.
\textbf{Symmetry preserving:} If two features play the exactly same role in the network, they should receive the same attribution. 



Since all prior prototype based models generate attribution maps for explaining ``where" the model looks at based on the final feature map $\mathbf{x}$, we consider the final prediction attributed to this feature map when discussing the above axioms. The process of calculating the logit for the class c in our method can be expressed as 
\begin{equation}
\text{Avg} (\mathbf{x}\mathbf{v}^T) = \text{Avg}(\mathbf{x}\Tilde{\mathbf{p}}_1^T+...+\mathbf{x}\Tilde{\mathbf{p}}_k^T),
\end{equation}
where $\text{Avg}$ is the global average pooling. In the prior prototype based model, this logit is calculated via
\begin{equation}
    h (\text{max}(g(\mathbf{x},\mathbf{p}_1')),...,\text{max}(g(\mathbf{x},\mathbf{p}_k'))),
\end{equation}
where $\mathbf{p}_i'$ are prototypes, $\text{max}$ is the global max pooling operation to obtain the similarity between a feature map and a prototype, $g$ is a distance function and $h$ indicates a linear combination. Most works use max pooling for the similarity score \cite{chen2019looks,wang2021interpretable,rymarczyk2022interpretable,huang2023evaluation,donnelly2022deformable}, except ProtoTree \cite{nauta2021neural} that uses a min pooling. A summary of the comparison using explainability axioms is presented in Table \ref{tab:xai_property}. Since the attributions yielded by our method directly sum up linearly to the final prediction, all four axioms are fulfilled. In contrast, due to the usage of max pooling, prior models break the \textit{sensitivity} axiom as some areas receiving attributions in the visualization $g(\mathbf{x},\mathbf{p}_i')$ may have no contribution to the final prediction. For the same reason, some similarities between specific features and prototypes do not contribute to the predicted logit, breaking the \textit{completeness} axiom. It's also straightforward that the max pooling breaks the \textit{linearity} axiom. 
The only axiom that prior baselines fulfill is the symmetry preserving axiom because the order of variables does not influence the value after the max pooling.

\begin{table}[htbp]
\vspace{-0.2cm}
\caption{Comparison with prior prototype based models using explainability axioms sensitivity (SNES.), completeness (COMPL.), linearity (LIN.) and symmetry-preserving (SYM.).}
\label{tab:xai_property}
\begin{center}
\begin{small}
\begin{sc}
\begin{tabular}{lccccr}
\toprule
Methods & Sens. & Compl. & Lin.  & Sym.\\
\midrule
Baselines &  $\times$  & $\times$ & $\times$ &   $\surd$\\
Ours  &$\surd$  &$\surd$&$\surd$& $\surd$\\
\bottomrule
\end{tabular}
\end{sc}
\end{small}
\end{center}
\vskip -0.1in
\end{table}

\subsection{Comparisons via Quantitative Metrics}
In this study, we compare our post-hoc part-prototype model with prior baseline models using interpretability and accuracy metrics quantitatively.

\textbf{Implementation details.} Following prior works, we use the benchmark dataset CUB-200-2011 \cite{wah2011caltech}, which consists of 200 classes of bird spieces for this comparison. Although there are results reported in prior work \cite{chen2019looks} as a baseline for trained original ResNet \cite{he2016deep}, VGG \cite{simonyan2014very}, DenseNet \cite{iandola2014densenet} models, we train these original models ourselves using a simple training scheme and describe the details as follows: we leverage the ImageNet \cite{deng2009imagenet} pretrained models as initialization for the backbone following \cite{chen2019looks}, and replace the classification head with randomly initialized linear classifier capable of classifying 200 classes to fit the dataset class number. We train 100 epochs with initial learning rate 0.001 and multiply it by 0.1 every 30 epochs. We use SGD as the optimizer with momentum 0.9 and weight decay $10^{-4}$. We only use horizontal flip with probability 0.5 as data augmentation. We train the whole network without specifically freezing any layer and apply simple cross entropy loss. Note that the above training scheme for training standard Res/VGG/DenseNet networks  is significantly simpler than prior prototype based models, which typically require adding an additional prototype layer with learnable parameters, iteratively freezing and fine-tuning certain layers, carefully setting different learning rates for different layers and a couple of special loss or distance function designs \cite{chen2019looks, wang2021interpretable,donnelly2022deformable,huang2023evaluation}. 

\begin{table*}[t]
\caption{Comprehensive evaluation of interpretability using consistency scores (CON.), stability scores (STA.) as well as the accuracy (ACC.) of part-prototype networks on the CUB-200-2011 \cite{wah2011caltech} dataset. The results are over five backbones pretrained in ImageNet \cite{deng2009imagenet}. Best in bold.}
\label{tab:cub_performance}
\vskip 0.15in
\begin{center}
\begin{small}
\begin{sc}
\begin{tabular}{l *{15}{p{0.021\textwidth}}} 
\toprule
& \multicolumn{3}{c}{ResNet34} & \multicolumn{3}{|c|}{ResNet152} &\multicolumn{3}{|c|}{VGG19}  &\multicolumn{3}{c}{Dense121} &\multicolumn{3}{|c}{Dense161} \\
\cline{2-16}
Methods & con. & sta.&acc. &con. & sta.&acc. &con. & sta.&acc. &con. & sta.&acc. &con. & sta.&acc. \\
\midrule
Tree. \cite{nauta2021neural} & 10.0 & 21.6& 70.1 & 16.4 &23.2&71.2 & 17.6& 19.8& 68.7& 21.5& 24.4& 73.2& 18.8& 28.9& 72.4\\
Net. \cite{chen2019looks} & 15.1&53.8&79.2&28.3&56.7&78.0&31.6&60.4&78.0&24.9&58.9&80.2&21.2&58.2&80.1\\
Pool. \cite{rymarczyk2022interpretable}   & 32.4&57.6&80.3&35.7&58.4&81.5&36.2&\textbf{62.7}&78.4&48.5&55.3&81.5&40.6&61.2&82.0\\
Deform. \cite{donnelly2022deformable} & 39.9&57.0&81.1&44.2&53.5&82.0&40.6&61.5&77.9&61.4&64.7&82.6&46.7&63.9&83.3 \\
TesNet \cite{wang2021interpretable} & 53.3&65.4&\textbf{82.8}&48.6&60.0&82.7&46.8&58.2&\textbf{81.4}&63.1&66.1&\textbf{84.8}&\textbf{62.2}&67.5&84.6\\
Eval. \cite{huang2023evaluation} & 52.9&66.3&82.3&50.7&65.7&83.8&44.6&56.9&80.6&52.9&59.1&83.4&56.3&61.5&84.7\\
\midrule
\textit{Ours} & \textbf{62.3}&\textbf{79.5}&82.3&\textbf{60.0}&\textbf{76.0}&\textbf{86.2}&\textbf{49.7}&56.9&81.2&\textbf{65.2}&\textbf{82.8}&84.1&60.0&\textbf{80.7}&\textbf{85.9}\\
\bottomrule
\end{tabular}
\end{sc}
\end{small}
\end{center}
\vskip -0.1in
\end{table*}

\textbf{Quantitative interpretability and accuracy comparisons.} The quality of the prototype is evaluated by the consistency and stability scores following \cite{huang2023evaluation}. The consistency score calculates the percentage of all learned prototypes that could consistently activate the semantically same areas (e.g., a bird head) across different images using keypoint annotations offered by the dataset. The stability score calculates the percentage of prototypes that activate the same areas after the images are disturbed by noises. We use the base version of ProtoEval \cite{huang2023evaluation} for a fair comparison with other prior works. Table \ref{tab:cub_performance} shows that in Res34, Res152, VGG19, Dense121, our method outperforms prior works in the consistency scores by +9.4 \%, +9.3\%, +5.1\%, +3.8\%, respectively.
In Res34, Res152, Dense121, Dense161 backbones, our stability score outperforms prior works by an even larger margin (+12\%,+11\%,+23.7\%,+19.2\%). Such simply trained models with no special design easily suppresses nearly all prior works in the prediction accuracy. The interpretability metrics in our method are calculated using $k=3$ prototypes for each class and using more prototypes would yield even larger advantages, as shown in the ablation study. We note that all possible fine-tuning tricks or specific loss designs for fine-grained image classification could easily be applied to further increase the accuracy. However, this is not our goal and we show that such a simply trained model could readily yield prototypes with very good qualities. 

\subsection{Analysis in the Large Scale Dataset ImageNet}

\paragraph{Performance comparisons.} None of the prior prototype based methods evaluates their models in this large scale dataset. Some work even directly indicates that their model may be computationally too heavy in large scale datasets \cite{ukai2022looks}. To test their scalability, 
we use the ImageNet pretrained ResNet34 as backbone and try to run the released code of four representative models, namely ProtopNet \cite{chen2019looks}, TesNet \cite{wang2021interpretable}, ProtoPool \cite{rymarczyk2022interpretable}, ProtoDeform \cite{donnelly2022deformable} by changing the class number and corresponding prototype numbers to observe how would the performance change after equipping the pretrained model with the part-prototype based global interpretability. The results are summarized in Table \ref{tab:imagenet_performance}. Both ProtoPool \cite{rymarczyk2022interpretable} and TesNet \cite{wang2021interpretable} are not runnable even with batch size 1 in a NVIDIA 3090 Ti with 24 GB memory, indicating their huge memory consumption in complex handling of prototypes when the class number is large. Other methods such as ProtopNet \cite{chen2019looks} and ProtoDeform \cite{donnelly2022deformable} exhibit a strong performance drop (-9.6\%, -11.7\%) compared to the original pre-trained ResNet34. These results indicate the good scalability of our post-hoc approach and its value in maintaining the accuracy in large scale datasets.

\begin{table}[htp]
\vspace{-0.2cm}
\caption{Performance in ImageNet of ProtopNet (Net) \cite{chen2019looks}, TesNet (Tes) \cite{wang2021interpretable}, ProtoPool (Pool) \cite{rymarczyk2022interpretable}, ProtoDeform (Deform) \cite{donnelly2022deformable} respectively. All methods either suffer from strong performance drop or can not scale to this large scale dataset. OOM indicates the out of memory error even with batch size 1 on a 24 GB GPU.}
\label{tab:imagenet_performance}
\vspace{-0.2cm}
\begin{center}
\begin{small}
\begin{sc}
\begin{tabular}{lccccccr}
\toprule
Methods & Net & Tes & Pool & Deform  & \emph{Ours} \\
\midrule
Accuracies & 65.5 & OOM & OOM &63.4&\textbf{75.1}  \\
\bottomrule
\end{tabular}
\end{sc}
\end{small}
\end{center}
\end{table}

\begin{figure}[h!]
	\centering
	\includegraphics[width=6cm]{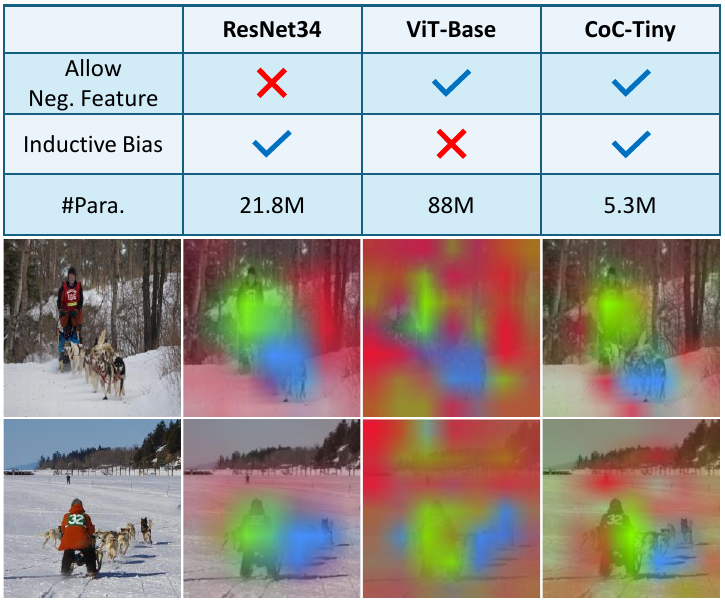}
 \caption{Comparison of 3 architectures and their prototypes in ImageNet \citep{deng2009imagenet}. We visualize the presence areas of 3 prototypes in 3 example images from the class "sled dog".}
	\label{fig:imagenet_protytpes}
 \vspace{-0.5cm}
\end{figure}

\paragraph{Qualitative comparisons of NMF prototypes in 3 architectures.}
CNN architectures \citep{he2016deep, simonyan2014very} typically use ReLU \citep{nair2010rectified} as the activation functions and naturally obtain non-negative features. However, more recent architectures such as ViT \citep{dosovitskiy2020image} or CoC \citep{ma2023image} use GeLU \citep{hendrycks2016gaussian} as activation functions and thus allow negative feature values. We note that there are variants of NMF such as semi-NMF and convex-NMF which could handle negative input values \citep{ding2008convex}. However, for consistency in the evaluation across different architectures, we simply set all negative features to zero and conduct NMF on extracted deep features. We evaluate multiple architectures including ResNet34 \citep{he2016deep}, ViT-base with patch size 32 \citep{dosovitskiy2020image} and Coc-tiny \citep{ma2023image}. A brief summary of different architectures is shown in the Figure \ref{fig:imagenet_protytpes}. The areas that the prototypes are present are visualized by different colors. It could be seen that even if we remove the negative feature values in ViT \cite{dosovitskiy2020image} and CoC \cite{ma2023image}, NMF still leads to reasonable parts based representation. This indicates that the non-negative feature values already capture the major representations of the corresponding image parts. Moreover, the areas of different prototypes from ViT are less separable, probably because the ViT architecture has less inductive bias. This may make the spatial relationship in the original image less maintained in the feature map.

\subsection{Ablation study}
In this section, we ablate our design choices under different number of prototypes to analyze the importance regarding the interpretability as well as the accuracy. 

\begin{figure}
    \centering
    \includegraphics[width=0.8\linewidth]{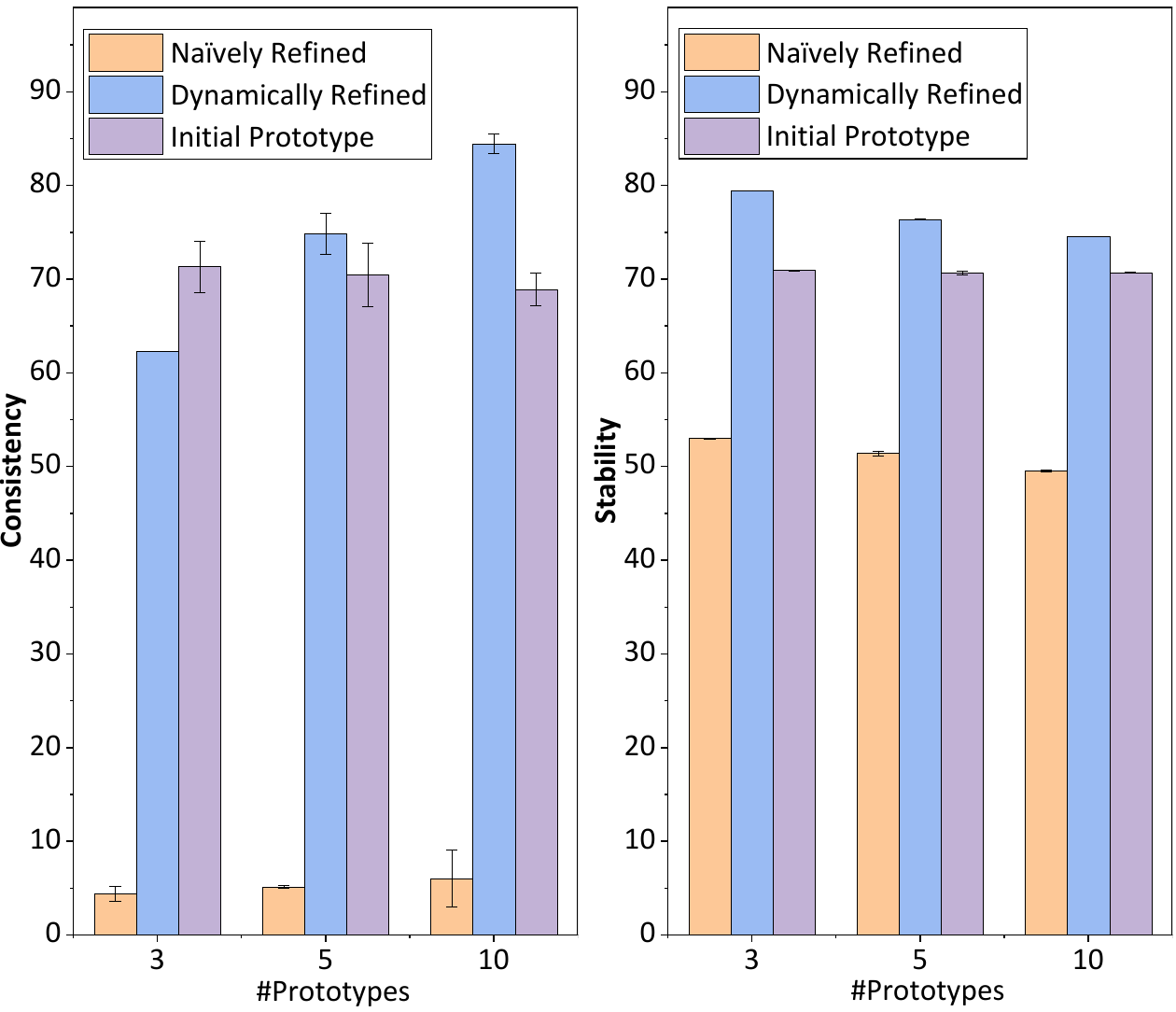}
    \caption{Interpretability benefits of our proposed prototype refinement strategy measured by prototype's consistency and stability scores. Our dynamic refinement strategy (blue) not only outperforms the naive distribution (orange) by a huge margin, but also surprisingly improves the initial prototypes (purple) in most cases.} 
    \vspace{-0.11cm}\label{fig:interpretability_constraint}
\end{figure}

\textbf{Importance of the interpretability constraints.}
We use the trained ResNet34 in CUB-200-2011 \cite{wah2011caltech} to conduct this study. As shown in Figure \ref{fig:interpretability_constraint}, our method outperforms the naive version of prototype refinement by a very large margin in all cases. Surprisingly, our method not only successfully maintain the interpretability of the initially discovered prototypes, but also significantly improves the prototype quality measured by consistency and stability scores in the case of $k=5$ and $k=10$. 

\textbf{Importance of the prototype refinement for accuracies.}
We conduct this study in the ImageNet \cite{deng2009imagenet} dataset using ResNet34 \cite{he2016deep}, ViT-Base \cite{dosovitskiy2020image}, CoC-Tiny \cite{ma2023image}. As could be seen in the Table \ref{tab:performance_net_nmf}, our refinement step plays an important role in guaranteeing the model's prediction accuracy, indicating the importance of equipping interpretable prototypes with discriminative power via our prototype refinement step. Note that the result 76.3 of ViT comes from averaged features of the last layer into the classification head for fair comparison. Using the ViT's classification token reaches 80.7 Top1 accuracy.

\begin{table}[h]
\caption{Performance comparison using initial prototypes and finally refined prototypes in ImageNet \cite{deng2009imagenet}.}
\label{tab:performance_net_nmf}
\begin{center}
\begin{small}
\begin{sc}
\begin{tabular}{lcccccr}
\toprule
& \multicolumn{3}{c}{Initial prototypes} & \multicolumn{3}{|c}{Refined prototypes}  \\
\cline{2-7}
\#Proto & Res & ViT&CoC & Res & ViT&CoC \\
\midrule
k=3 & 16.4 & 50.8 & 45.1&75.1 & 76.3 & 71.9\\
k=5 & 20.4 & 52.1 & 49.3&75.1 & 76.3 & 71.9\\
k=10   & 28.4 & 54.8 & 54.3&75.1 & 76.3 & 71.9\\
\bottomrule
\end{tabular}
\end{sc}
\end{small}
\end{center}
\end{table}

\begin{figure}
    \centering
    \includegraphics[width=0.7\linewidth]{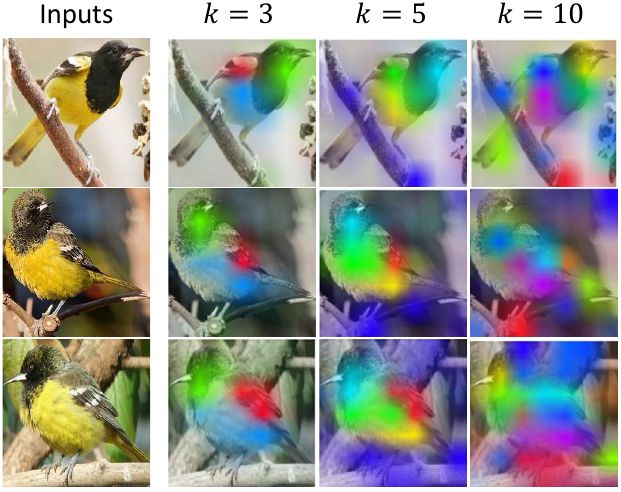}
    \caption{Visualizations of discovered prototypes using different prototype number $k$ in CUB. Setting $k=3$ discovers head, belly and wing consistently across different images. Increasing this number to 5 can additionally discover prototypes such as legs (deep blue) and breast (green color). $k=10$ discovers even more fine-grained interpretable prototypes such as throat and tail.}
    \label{fig:different_k}
\end{figure}

\textbf{Visualizations of different number of prototypes.} 
Figure \ref{fig:different_k} shows setting $k=3$ successfully discovers part-prototypes corresponding to head, wing and belly. When $k=5$, prototypes of breast and legs are additionally discovered. Even $k=10$ discovers surprisingly meaningful object parts (e.g., head, wing, throat, back, breast, belly, legs, tail, etc). The flexibility of obtaining various number of part-prototypes is an advantage of our post-hoc method for rich explanations. In contrast, prior methods require re-training the whole model when varying the prototype number. 

\subsection{Limitations}
As a post-hoc method, the capability of our method in obtaining part-prototypes is limited by the trained black box model. For example, if the trained model can not extract distinct features for certain objects, it is also hard for our algorithm to discover corresponding part-prototypes.

\section{Conclusion}
In this work, we aim to answer: can we construct a model that explains both ``where" the model looks at and ``what" the model looks for in a post-hoc manner to guarantee the performance? To this end, we propose the first post-hoc part-prototype network via a novel prototype discovery and refinement strategy which precisely reconstructs the trained classification head. Besides guaranteeing the performance, our method is even more faithful when explaining ``where" and yields more consistent and stable prototypes when explaining ``what" compared to prior works. This work indicates the value of exploring existing feature space for interpretability and opens up a new paradigm to design part-prototype networks in a post-hoc manner.

\section*{Acknowledgements}
The authors would like to thank all anonymous reviewers for insightful suggestions. This work was supported by the Hong Kong Innovation and Technology Fund (Project No. MHP/002/22) and HKUST (Project No. FS111).

\section*{Impact Statement}
Constructing a part-prototype network helps human to better understand how deep neural networks make decisions. Our post-hoc method significantly simplifies the construction process and yields interpretable prototypes of even better qualities while guaranteeing the accuracy. Our method has the potential to increase transparency and accountability in AI applications, which is crucial for building trust with users and stakeholders. We believe that our method represents a significant step forward in the development of responsible and transparent AI systems.

\nocite{langley00}


\begin{thebibliography}{57}
\providecommand{\natexlab}[1]{#1}
\providecommand{\url}[1]{\texttt{#1}}
\expandafter\ifx\csname urlstyle\endcsname\relax
  \providecommand{\doi}[1]{doi: #1}\else
  \providecommand{\doi}{doi: \begingroup \urlstyle{rm}\Url}\fi

\bibitem[Alvarez~Melis \& Jaakkola(2018)Alvarez~Melis and Jaakkola]{alvarez2018towards}
Alvarez~Melis, D. and Jaakkola, T.
\newblock Towards robust interpretability with self-explaining neural networks.
\newblock \emph{Advances in neural information processing systems}, 31, 2018.

\bibitem[Biederman(1987)]{biederman1987recognition}
Biederman, I.
\newblock Recognition-by-components: a theory of human image understanding.
\newblock \emph{Psychological review}, 94\penalty0 (2):\penalty0 115, 1987.

\bibitem[Carmichael et~al.(2024)Carmichael, Lohit, Cherian, Jones, and Scheirer]{carmichael2024pixel}
Carmichael, Z., Lohit, S., Cherian, A., Jones, M.~J., and Scheirer, W.~J.
\newblock Pixel-grounded prototypical part networks.
\newblock In \emph{Proceedings of the IEEE/CVF Winter Conference on Applications of Computer Vision}, pp.\  4768--4779, 2024.

\bibitem[Chen et~al.(2019)Chen, Li, Tao, Barnett, Rudin, and Su]{chen2019looks}
Chen, C., Li, O., Tao, D., Barnett, A., Rudin, C., and Su, J.~K.
\newblock This looks like that: deep learning for interpretable image recognition.
\newblock \emph{Advances in neural information processing systems}, 32, 2019.

\bibitem[Colin et~al.(2022)Colin, Fel, Cad{\`e}ne, and Serre]{colin2022cannot}
Colin, J., Fel, T., Cad{\`e}ne, R., and Serre, T.
\newblock What i cannot predict, i do not understand: A human-centered evaluation framework for explainability methods.
\newblock \emph{Advances in Neural Information Processing Systems}, 35:\penalty0 2832--2845, 2022.

\bibitem[Collins et~al.(2018)Collins, Achanta, and Susstrunk]{collins2018deep}
Collins, E., Achanta, R., and Susstrunk, S.
\newblock Deep feature factorization for concept discovery.
\newblock In \emph{Proceedings of the European Conference on Computer Vision (ECCV)}, pp.\  336--352, 2018.

\bibitem[Deng et~al.(2009)Deng, Dong, Socher, Li, Li, and Fei-Fei]{deng2009imagenet}
Deng, J., Dong, W., Socher, R., Li, L.-J., Li, K., and Fei-Fei, L.
\newblock Imagenet: A large-scale hierarchical image database.
\newblock In \emph{2009 IEEE conference on computer vision and pattern recognition}, pp.\  248--255. Ieee, 2009.

\bibitem[Diamond \& Boyd(2016)Diamond and Boyd]{diamond2016cvxpy}
Diamond, S. and Boyd, S.
\newblock {CVXPY}: {A} {P}ython-embedded modeling language for convex optimization.
\newblock \emph{Journal of Machine Learning Research}, 17\penalty0 (83):\penalty0 1--5, 2016.

\bibitem[Ding et~al.(2008)Ding, Li, and Jordan]{ding2008convex}
Ding, C.~H., Li, T., and Jordan, M.~I.
\newblock Convex and semi-nonnegative matrix factorizations.
\newblock \emph{IEEE transactions on pattern analysis and machine intelligence}, 32\penalty0 (1):\penalty0 45--55, 2008.

\bibitem[Donnelly et~al.(2022)Donnelly, Barnett, and Chen]{donnelly2022deformable}
Donnelly, J., Barnett, A.~J., and Chen, C.
\newblock Deformable protopnet: An interpretable image classifier using deformable prototypes.
\newblock In \emph{Proceedings of the IEEE/CVF Conference on Computer Vision and Pattern Recognition}, pp.\  10265--10275, 2022.

\bibitem[Dosovitskiy et~al.(2020)Dosovitskiy, Beyer, Kolesnikov, Weissenborn, Zhai, Unterthiner, Dehghani, Minderer, Heigold, Gelly, et~al.]{dosovitskiy2020image}
Dosovitskiy, A., Beyer, L., Kolesnikov, A., Weissenborn, D., Zhai, X., Unterthiner, T., Dehghani, M., Minderer, M., Heigold, G., Gelly, S., et~al.
\newblock An image is worth 16x16 words: Transformers for image recognition at scale.
\newblock \emph{arXiv preprint arXiv:2010.11929}, 2020.

\bibitem[Fel et~al.(2023)Fel, Picard, Bethune, Boissin, Vigouroux, Colin, Cad{\`e}ne, and Serre]{fel2023craft}
Fel, T., Picard, A., Bethune, L., Boissin, T., Vigouroux, D., Colin, J., Cad{\`e}ne, R., and Serre, T.
\newblock Craft: Concept recursive activation factorization for explainability.
\newblock In \emph{Proceedings of the IEEE/CVF Conference on Computer Vision and Pattern Recognition}, pp.\  2711--2721, 2023.

\bibitem[Frey \& Pimentel(1978)Frey and Pimentel]{frey1978principal}
Frey, D. and Pimentel, R.
\newblock Principal component analysis and factor analysis.
\newblock 1978.

\bibitem[Girshick(2015)]{girshick2015fast}
Girshick, R.
\newblock Fast r-cnn.
\newblock In \emph{Proceedings of the IEEE international conference on computer vision}, pp.\  1440--1448, 2015.

\bibitem[Gray(1984)]{gray1984vector}
Gray, R.
\newblock Vector quantization.
\newblock \emph{IEEE Assp Magazine}, 1\penalty0 (2):\penalty0 4--29, 1984.

\bibitem[He et~al.(2016)He, Zhang, Ren, and Sun]{he2016deep}
He, K., Zhang, X., Ren, S., and Sun, J.
\newblock Deep residual learning for image recognition.
\newblock In \emph{Proceedings of the IEEE conference on computer vision and pattern recognition}, pp.\  770--778, 2016.

\bibitem[Hendrycks \& Gimpel(2016)Hendrycks and Gimpel]{hendrycks2016gaussian}
Hendrycks, D. and Gimpel, K.
\newblock Gaussian error linear units (gelus).
\newblock \emph{arXiv preprint arXiv:1606.08415}, 2016.

\bibitem[Huang et~al.(2023)Huang, Xue, Huang, Zhang, Song, Jing, and Song]{huang2023evaluation}
Huang, Q., Xue, M., Huang, W., Zhang, H., Song, J., Jing, Y., and Song, M.
\newblock Evaluation and improvement of interpretability for self-explainable part-prototype networks.
\newblock In \emph{Proceedings of the IEEE/CVF International Conference on Computer Vision}, pp.\  2011--2020, 2023.

\bibitem[Hyv{\"a}rinen \& Oja(2000)Hyv{\"a}rinen and Oja]{hyvarinen2000independent}
Hyv{\"a}rinen, A. and Oja, E.
\newblock Independent component analysis: algorithms and applications.
\newblock \emph{Neural networks}, 13\penalty0 (4-5):\penalty0 411--430, 2000.

\bibitem[Iandola et~al.(2014)Iandola, Moskewicz, Karayev, Girshick, Darrell, and Keutzer]{iandola2014densenet}
Iandola, F., Moskewicz, M., Karayev, S., Girshick, R., Darrell, T., and Keutzer, K.
\newblock Densenet: Implementing efficient convnet descriptor pyramids.
\newblock \emph{arXiv preprint arXiv:1404.1869}, 2014.

\bibitem[Kapishnikov et~al.(2021)Kapishnikov, Venugopalan, Avci, Wedin, Terry, and Bolukbasi]{kapishnikov2021guided}
Kapishnikov, A., Venugopalan, S., Avci, B., Wedin, B., Terry, M., and Bolukbasi, T.
\newblock Guided integrated gradients: An adaptive path method for removing noise.
\newblock In \emph{Proceedings of the IEEE/CVF conference on computer vision and pattern recognition}, pp.\  5050--5058, 2021.

\bibitem[Khosla et~al.(2011)Khosla, Jayadevaprakash, Yao, and Fei-Fei]{KhoslaYaoJayadevaprakashFeiFei_FGVC2011}
Khosla, A., Jayadevaprakash, N., Yao, B., and Fei-Fei, L.
\newblock Novel dataset for fine-grained image categorization.
\newblock In \emph{First Workshop on Fine-Grained Visual Categorization, IEEE Conference on Computer Vision and Pattern Recognition}, Colorado Springs, CO, June 2011.

\bibitem[Kim et~al.(2022)Kim, Meister, Ramaswamy, Fong, and Russakovsky]{kim2022hive}
Kim, S.~S., Meister, N., Ramaswamy, V.~V., Fong, R., and Russakovsky, O.
\newblock Hive: Evaluating the human interpretability of visual explanations.
\newblock In \emph{European Conference on Computer Vision}, pp.\  280--298. Springer, 2022.

\bibitem[Krause et~al.(2013)Krause, Stark, Deng, and Fei-Fei]{krause20133d}
Krause, J., Stark, M., Deng, J., and Fei-Fei, L.
\newblock 3d object representations for fine-grained categorization.
\newblock In \emph{Proceedings of the IEEE international conference on computer vision workshops}, pp.\  554--561, 2013.

\bibitem[Lee \& Seung(2000)Lee and Seung]{lee2000algorithms}
Lee, D. and Seung, H.~S.
\newblock Algorithms for non-negative matrix factorization.
\newblock \emph{Advances in neural information processing systems}, 13, 2000.

\bibitem[Lee \& Seung(1999)Lee and Seung]{lee1999learning}
Lee, D.~D. and Seung, H.~S.
\newblock Learning the parts of objects by non-negative matrix factorization.
\newblock \emph{Nature}, 401\penalty0 (6755):\penalty0 788--791, 1999.

\bibitem[Lundberg \& Lee(2017)Lundberg and Lee]{lundberg2017unified}
Lundberg, S.~M. and Lee, S.-I.
\newblock A unified approach to interpreting model predictions.
\newblock \emph{Advances in neural information processing systems}, 30, 2017.

\bibitem[Ma et~al.(2023{\natexlab{a}})Ma, Zhao, Chen, and Rudin]{ma2023looks}
Ma, C., Zhao, B., Chen, C., and Rudin, C.
\newblock This looks like those: Illuminating prototypical concepts using multiple visualizations.
\newblock \emph{arXiv preprint arXiv:2310.18589}, 2023{\natexlab{a}}.

\bibitem[Ma et~al.(2023{\natexlab{b}})Ma, Zhou, Wang, Qin, Sun, Liu, and Fu]{ma2023image}
Ma, X., Zhou, Y., Wang, H., Qin, C., Sun, B., Liu, C., and Fu, Y.
\newblock Image as set of points.
\newblock In \emph{The Eleventh International Conference on Learning Representations}, 2023{\natexlab{b}}.
\newblock URL \url{https://openreview.net/forum?id=awnvqZja69}.

\bibitem[Nair \& Hinton(2010)Nair and Hinton]{nair2010rectified}
Nair, V. and Hinton, G.~E.
\newblock Rectified linear units improve restricted boltzmann machines.
\newblock In \emph{Proceedings of the 27th international conference on machine learning (ICML-10)}, pp.\  807--814, 2010.

\bibitem[Nauta et~al.(2021)Nauta, Van~Bree, and Seifert]{nauta2021neural}
Nauta, M., Van~Bree, R., and Seifert, C.
\newblock Neural prototype trees for interpretable fine-grained image recognition.
\newblock In \emph{Proceedings of the IEEE/CVF Conference on Computer Vision and Pattern Recognition}, pp.\  14933--14943, 2021.

\bibitem[Nelder \& Mead(1965)Nelder and Mead]{nelder1965simplex}
Nelder, J.~A. and Mead, R.
\newblock A simplex method for function minimization.
\newblock \emph{The computer journal}, 7\penalty0 (4):\penalty0 308--313, 1965.

\bibitem[Ramaswamy et~al.(2023)Ramaswamy, Kim, Fong, and Russakovsky]{ramaswamy2023overlooked}
Ramaswamy, V.~V., Kim, S.~S., Fong, R., and Russakovsky, O.
\newblock Overlooked factors in concept-based explanations: Dataset choice, concept learnability, and human capability.
\newblock In \emph{Proceedings of the IEEE/CVF Conference on Computer Vision and Pattern Recognition}, pp.\  10932--10941, 2023.

\bibitem[Ribeiro et~al.(2016)Ribeiro, Singh, and Guestrin]{ribeiro2016should}
Ribeiro, M.~T., Singh, S., and Guestrin, C.
\newblock " why should i trust you?" explaining the predictions of any classifier.
\newblock In \emph{Proceedings of the 22nd ACM SIGKDD international conference on knowledge discovery and data mining}, pp.\  1135--1144, 2016.

\bibitem[Ronneberger et~al.(2015)Ronneberger, Fischer, and Brox]{ronneberger2015u}
Ronneberger, O., Fischer, P., and Brox, T.
\newblock U-net: Convolutional networks for biomedical image segmentation.
\newblock In \emph{Medical Image Computing and Computer-Assisted Intervention--MICCAI 2015: 18th International Conference, Munich, Germany, October 5-9, 2015, Proceedings, Part III 18}, pp.\  234--241. Springer, 2015.

\bibitem[Rudin(2019)]{rudin2019stop}
Rudin, C.
\newblock Stop explaining black box machine learning models for high stakes decisions and use interpretable models instead.
\newblock \emph{Nature machine intelligence}, 1\penalty0 (5):\penalty0 206--215, 2019.

\bibitem[Rymarczyk et~al.(2021)Rymarczyk, Struski, Tabor, and Zieli{\'n}ski]{rymarczyk2021protopshare}
Rymarczyk, D., Struski, {\L}., Tabor, J., and Zieli{\'n}ski, B.
\newblock Protopshare: Prototypical parts sharing for similarity discovery in interpretable image classification.
\newblock In \emph{Proceedings of the 27th ACM SIGKDD Conference on Knowledge Discovery \& Data Mining}, pp.\  1420--1430, 2021.

\bibitem[Rymarczyk et~al.(2022)Rymarczyk, Struski, G{\'o}rszczak, Lewandowska, Tabor, and Zieli{\'n}ski]{rymarczyk2022interpretable}
Rymarczyk, D., Struski, {\L}., G{\'o}rszczak, M., Lewandowska, K., Tabor, J., and Zieli{\'n}ski, B.
\newblock Interpretable image classification with differentiable prototypes assignment.
\newblock In \emph{Computer Vision--ECCV 2022: 17th European Conference, Tel Aviv, Israel, October 23--27, 2022, Proceedings, Part XII}, pp.\  351--368. Springer, 2022.

\bibitem[Schwalbe \& Finzel(2023)Schwalbe and Finzel]{schwalbe2023comprehensive}
Schwalbe, G. and Finzel, B.
\newblock A comprehensive taxonomy for explainable artificial intelligence: a systematic survey of surveys on methods and concepts.
\newblock \emph{Data Mining and Knowledge Discovery}, pp.\  1--59, 2023.

\bibitem[Selvaraju et~al.(2017)Selvaraju, Cogswell, Das, Vedantam, Parikh, and Batra]{selvaraju2017grad}
Selvaraju, R.~R., Cogswell, M., Das, A., Vedantam, R., Parikh, D., and Batra, D.
\newblock Grad-cam: Visual explanations from deep networks via gradient-based localization.
\newblock In \emph{Proceedings of the IEEE international conference on computer vision}, pp.\  618--626, 2017.

\bibitem[Simonyan \& Zisserman(2014)Simonyan and Zisserman]{simonyan2014very}
Simonyan, K. and Zisserman, A.
\newblock Very deep convolutional networks for large-scale image recognition.
\newblock \emph{arXiv preprint arXiv:1409.1556}, 2014.

\bibitem[Sundararajan et~al.(2017)Sundararajan, Taly, and Yan]{sundararajan2017axiomatic}
Sundararajan, M., Taly, A., and Yan, Q.
\newblock Axiomatic attribution for deep networks.
\newblock In \emph{International conference on machine learning}, pp.\  3319--3328. PMLR, 2017.

\bibitem[Tenenbaum \& Freeman(1996)Tenenbaum and Freeman]{tenenbaum1996separating}
Tenenbaum, J. and Freeman, W.
\newblock Separating style and content.
\newblock \emph{Advances in neural information processing systems}, 9, 1996.

\bibitem[Tenenbaum et~al.(2000)Tenenbaum, Silva, and Langford]{tenenbaum2000global}
Tenenbaum, J.~B., Silva, V.~d., and Langford, J.~C.
\newblock A global geometric framework for nonlinear dimensionality reduction.
\newblock \emph{science}, 290\penalty0 (5500):\penalty0 2319--2323, 2000.

\bibitem[Ukai et~al.(2022)Ukai, Hirakawa, Yamashita, and Fujiyoshi]{ukai2022looks}
Ukai, Y., Hirakawa, T., Yamashita, T., and Fujiyoshi, H.
\newblock This looks like it rather than that: Protoknn for similarity-based classifiers.
\newblock In \emph{The Eleventh International Conference on Learning Representations}, 2022.

\bibitem[Virtanen et~al.(2020)Virtanen, Gommers, Oliphant, Haberland, Reddy, Cournapeau, Burovski, Peterson, Weckesser, Bright, {van der Walt}, Brett, Wilson, Millman, Mayorov, Nelson, Jones, Kern, Larson, Carey, Polat, Feng, Moore, {VanderPlas}, Laxalde, Perktold, Cimrman, Henriksen, Quintero, Harris, Archibald, Ribeiro, Pedregosa, {van Mulbregt}, and {SciPy 1.0 Contributors}]{2020SciPy-NMeth}
Virtanen, P., Gommers, R., Oliphant, T.~E., Haberland, M., Reddy, T., Cournapeau, D., Burovski, E., Peterson, P., Weckesser, W., Bright, J., {van der Walt}, S.~J., Brett, M., Wilson, J., Millman, K.~J., Mayorov, N., Nelson, A. R.~J., Jones, E., Kern, R., Larson, E., Carey, C.~J., Polat, {\.I}., Feng, Y., Moore, E.~W., {VanderPlas}, J., Laxalde, D., Perktold, J., Cimrman, R., Henriksen, I., Quintero, E.~A., Harris, C.~R., Archibald, A.~M., Ribeiro, A.~H., Pedregosa, F., {van Mulbregt}, P., and {SciPy 1.0 Contributors}.
\newblock {{SciPy} 1.0: Fundamental Algorithms for Scientific Computing in Python}.
\newblock \emph{Nature Methods}, 17:\penalty0 261--272, 2020.
\newblock \doi{10.1038/s41592-019-0686-2}.

\bibitem[Wachsmuth et~al.(1994)Wachsmuth, Oram, and Perrett]{wachsmuth1994recognition}
Wachsmuth, E., Oram, M., and Perrett, D.
\newblock Recognition of objects and their component parts: responses of single units in the temporal cortex of the macaque.
\newblock \emph{Cerebral Cortex}, 4\penalty0 (5):\penalty0 509--522, 1994.

\bibitem[Wah et~al.(2011)Wah, Branson, Welinder, Perona, and Belongie]{wah2011caltech}
Wah, C., Branson, S., Welinder, P., Perona, P., and Belongie, S.
\newblock The caltech-ucsd birds-200-2011 dataset.
\newblock 2011.

\bibitem[Wang et~al.(2023)Wang, Liu, Chen, Liu, Tian, McCarthy, Frazer, and Carneiro]{wang2023learning}
Wang, C., Liu, Y., Chen, Y., Liu, F., Tian, Y., McCarthy, D., Frazer, H., and Carneiro, G.
\newblock Learning support and trivial prototypes for interpretable image classification.
\newblock In \emph{Proceedings of the IEEE/CVF International Conference on Computer Vision}, pp.\  2062--2072, 2023.

\bibitem[Wang et~al.(2021)Wang, Liu, Wang, and Jing]{wang2021interpretable}
Wang, J., Liu, H., Wang, X., and Jing, L.
\newblock Interpretable image recognition by constructing transparent embedding space.
\newblock In \emph{Proceedings of the IEEE/CVF International Conference on Computer Vision}, pp.\  895--904, 2021.

\bibitem[Wang \& Zhang(2012)Wang and Zhang]{wang2012nonnegative}
Wang, Y.-X. and Zhang, Y.-J.
\newblock Nonnegative matrix factorization: A comprehensive review.
\newblock \emph{IEEE Transactions on knowledge and data engineering}, 25\penalty0 (6):\penalty0 1336--1353, 2012.

\bibitem[Wolf et~al.(2023)Wolf, Bongratz, Rickmann, P{\"o}lsterl, and Wachinger]{wolf2023keep}
Wolf, T.~N., Bongratz, F., Rickmann, A.-M., P{\"o}lsterl, S., and Wachinger, C.
\newblock Keep the faith: Faithful explanations in convolutional neural networks for case-based reasoning.
\newblock \emph{arXiv preprint arXiv:2312.09783}, 2023.

\bibitem[Yang et~al.(2023)Yang, Wang, and Bilgic]{yang2023idgi}
Yang, R., Wang, B., and Bilgic, M.
\newblock Idgi: A framework to eliminate explanation noise from integrated gradients.
\newblock In \emph{Proceedings of the IEEE/CVF Conference on Computer Vision and Pattern Recognition}, pp.\  23725--23734, 2023.

\bibitem[Zeiler \& Fergus(2014)Zeiler and Fergus]{zeiler2014visualizing}
Zeiler, M.~D. and Fergus, R.
\newblock Visualizing and understanding convolutional networks.
\newblock In \emph{Computer Vision--ECCV 2014: 13th European Conference, Zurich, Switzerland, September 6-12, 2014, Proceedings, Part I 13}, pp.\  818--833. Springer, 2014.

\bibitem[Zhang et~al.(2018)Zhang, Bargal, Lin, Brandt, Shen, and Sclaroff]{zhang2018top}
Zhang, J., Bargal, S.~A., Lin, Z., Brandt, J., Shen, X., and Sclaroff, S.
\newblock Top-down neural attention by excitation backprop.
\newblock \emph{International Journal of Computer Vision}, 126\penalty0 (10):\penalty0 1084--1102, 2018.

\bibitem[Zhou et~al.(2018)Zhou, Sun, Bau, and Torralba]{zhou2018interpretable}
Zhou, B., Sun, Y., Bau, D., and Torralba, A.
\newblock Interpretable basis decomposition for visual explanation.
\newblock In \emph{Proceedings of the European Conference on Computer Vision (ECCV)}, pp.\  119--134, 2018.

\bibitem[Zhou \& Troyanskaya(2015)Zhou and Troyanskaya]{zhou2015predicting}
Zhou, J. and Troyanskaya, O.~G.
\newblock Predicting effects of noncoding variants with deep learning--based sequence model.
\newblock \emph{Nature methods}, 12\penalty0 (10):\penalty0 931--934, 2015.

\end{thebibliography}

\bibliographystyle{icml2024}

\newpage
\appendix 
\onecolumn
\section{Case Study on Test-time Intervention}
\begin{figure}
    \centering
    \includegraphics[width=0.9\linewidth]{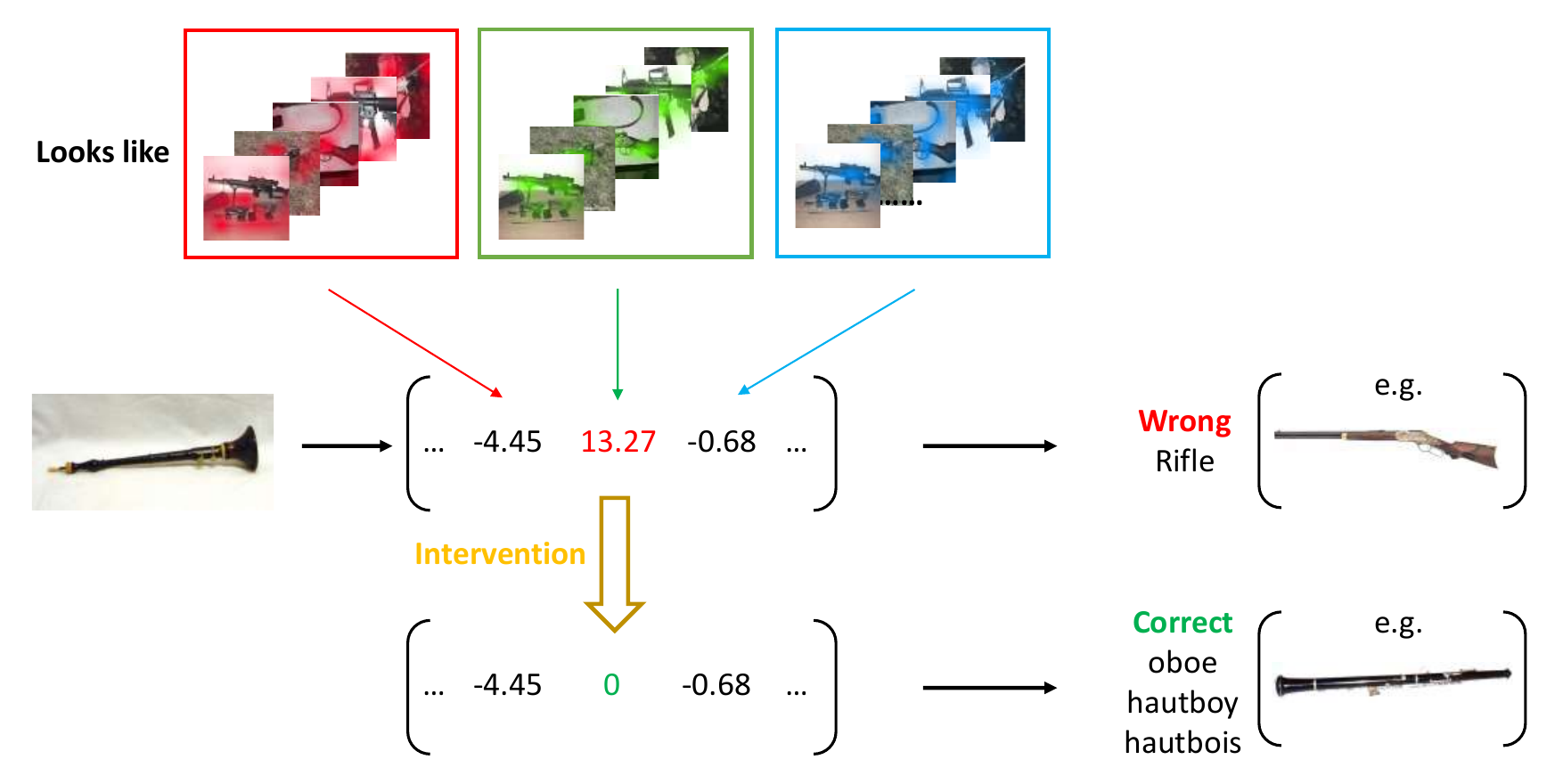}
    \caption{Example of test-time intervention: setting the contribution of the green prototype to zero corrects the model's prediction.}
    \label{fig:intervention}
\end{figure}

One of the major advantages of an interpretable model is enabling the human to correct the wrong predictions. In this section, we demonstrate a case study via Figure \ref{fig:intervention}. The initial prototypes discovered by NMF could be understood as follows: (1) The red prototype corresponds more to the background. (2) The green prototype is more in the shape of a long thin pipe, which confuses the model when the input is an instrument with the similar shape. (3) The blue prototype corresponds more to a trigger. So a human expert could intervene the green prototype by setting the contribution
from this prototype to zero and the model can now correctly predict the input image to obe, hautboy, hautbois.

\section{Accuracy Comparisons in Stanford Cars \& Dogs}
The Stanford Cars \cite{krause20133d} and Stanford Dogs \cite{KhoslaYaoJayadevaprakashFeiFei_FGVC2011} datasets are both for fine-grained image classification tasks. We show in Table \ref{tab:cars} and Table \ref{tab:dogs} that our model still outperforms prior works in the accuracy. All results are based on full images for a fair comparison.

\paragraph{Stanford Cars}
We follow the same data augmentations as ProtoTree \cite{nauta2021neural} and use the same training schedule as we do for the CUB-200-2011 \cite{wah2011caltech} dataset. We do not include any additional loss function design except the standard cross entropy and do not apply any fine-tuning tricks as prior existing prototype based models. The table \ref{tab:cars} indicates that such simply trained black box models still achieve a higher accuracy in 2 of the 3 backbones compared to prior state-of-the-art carefully engineered models in this dataset, further justifying the benefits of our post-hoc approach. 


\begin{table}[htbp]
\caption{Comparison with prior prototype based models under different backbones in the Stanford Cars \cite{krause20133d} dataset.}
\label{tab:cars}
\begin{center}
\begin{small}
\begin{sc}
\begin{tabular}{lccccr}
\toprule
Methods & VGG19 & Res34 & Res152 \\
\midrule
Tree \cite{nauta2021neural} & - & \textbf{86.6} & - \\
Deform \cite{donnelly2022deformable} &- & - & 86.5\\
ST \cite{wang2023learning} & - & - & 85.3\\
Support ST \cite{wang2023learning} & - & - & 87.3\\
PIXPNET\cite{carmichael2024pixel} &  86.4  & - & - \\
Ours  & \textbf{88.2}  & 85.5 & \textbf{89.7} \\
\bottomrule
\end{tabular}
\end{sc}
\end{small}
\end{center}
\end{table}

\paragraph{Stanford Dogs}
Since Deformable ProtopNet \cite{chen2019looks} does not release how they augment the data, we use simple random resized crop and random horizontal flip as the augmentation. The training scheme is the same as done for the CUB \cite{wah2011caltech} and Stanford Cars \cite{krause20133d} datasets. The accuracy comparisons are summarized in the following Table \ref{tab:dogs}. The black box models outperform prior works in all backbones. 

\begin{table}[htbp]
\caption{Comparison with prior prototype based models under different backbones in the Stanford Dogs \cite{KhoslaYaoJayadevaprakashFeiFei_FGVC2011} dataset.}
\label{tab:dogs}
\begin{center}
\begin{small}
\begin{sc}
\begin{tabular}{lccccr}
\toprule
Methods & VGG19 & Res152 & Dense161 \\
\midrule
ProtopNet \cite{chen2019looks} &  73.6  & 76.2 & 77.3 \\
Deform \cite{donnelly2022deformable} & 77.9 & 86.5 & 83.7\\
Ours  & \textbf{82.4}  & \textbf{87.2} & \textbf{85.1} \\
\bottomrule
\end{tabular}
\end{sc}
\end{small}
\end{center}
\end{table}




\section{More Implementation Details}
For NMF optimization, we use the update rules introduced in \citep{lee2000algorithms} and leverage the public code released by \cite{collins2018deep}.

For the prototype scaling via convex optimization, we use the cvxpy \cite{diamond2016cvxpy} package.

For optimization based residual parameter distribution subject to interpretability constraint, we only optimize $k-1$ components, and set the $k^{th}$ component as $\mathbf{R}-\sum_{i=1}^{k-1}\mathbf{r}_i$ to guarantee that all components sum up to $\mathbf{R}$ during the optimization. We use the SciPy \cite{2020SciPy-NMeth} package for this optimization.

\section{More Visualizations}
\begin{figure}
    \centering
    \includegraphics[width=0.7\linewidth]{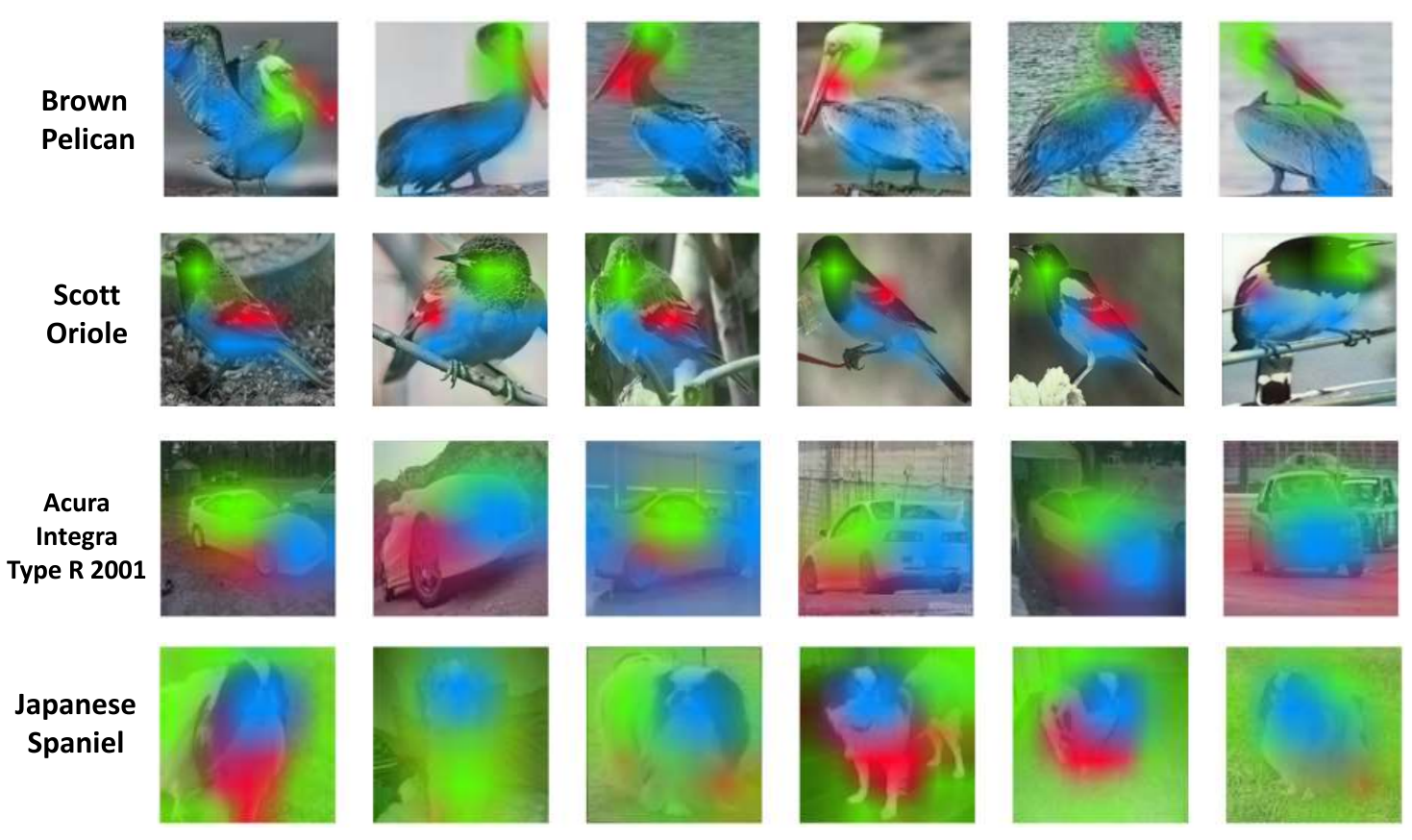}
    \caption{Visualization of prototypes in different classes. In each row, different colors indicate the activation areas of different prototypes. The prototypes discovered can consistently activate same semantically meaningful areas across images. This figure shows the case of $k=3$ prototypes in each class.}
    \label{fig:more_visualizations}
\end{figure}

In Figure \ref{fig:more_visualizations}, we show our discovered prototypes in each class can consistently activate the semantically same areas across different images in CUB \cite{wah2011caltech}, Cars \cite{krause20133d}, Dogs \cite{KhoslaYaoJayadevaprakashFeiFei_FGVC2011} respectively.

\end{document}